\documentclass{isprs} % isprs class modified 23-04-2019 (Dennis Wittich)
\usepackage{subcaption}
\usepackage{graphicx}
\usepackage{setspace}
\usepackage{geometry} % added 27-02-2014 Markus Englich
\usepackage{epstopdf}
\usepackage{natbib}
\usepackage{booktabs}
\usepackage{tabularx}
\usepackage{multirow}
\usepackage{adjustbox}
\usepackage{makecell}
\usepackage{float}

\usepackage[labelsep=period]{caption}  % added 14-04-2016 Markus Englich - Recommendation by Sebastian Brocks
\usepackage[british]{babel} 
\usepackage[hang]{footmisc}
\usepackage{siunitx}

\usepackage[%
  breaklinks=true,
  letterpaper=true,
  colorlinks,
  bookmarks=false,
]{hyperref}
\hypersetup{linkcolor={black}, urlcolor={blue}, citecolor={black}, colorlinks=true}

\usepackage[%
xindy,
nohypertypes=acronym,
]{glossaries}
\glsdisablehyper

\usepackage[capitalize]{cleveref}
\crefname{section}{Sec.}{Secs.}
\Crefname{section}{Section}{Sections}
\Crefname{table}{Table}{Tables}
\crefname{table}{Tab.}{Tabs.}

\newacronym{gl:AID}{AID}{Aerial Image Dataset} 
\newacronym{gl:CBAM}{CBAM}{Convolutional Block Attention Module}
\newacronym{gl:CNN}{CNN}{Convolutional Neural Network}
\newacronym{gl:MAE}{MAE}{Masked Autoencoder}
\newacronym{gl:SimCLR}{SimCLR}{Simple Framework for Contrastive Learning of Visual Representations}
\newacronym{gl:SwAV}{SwAV}{Swapping Assignments between Views}

\begin{document}

\title{Efficient Building Roof Type Classification: A Domain-Specific Self-Supervised Approach}
\date{}

% KAO: Remove extra spacing

% Anonymous submissions, authors' names should not be visible
\author{
 Guneet Mutreja\textsuperscript{1}, Ksenia Bittner\textsuperscript{1} }
% \author{***** (for review, names must be rendered anonymous)}

% KAO: Remove extra newline
% Anonymous submissions, authors' affiliations should not be visible
\address{\textsuperscript{1}Remote Sensing Technology Institute, German Aerospace Center (DLR), We\ss{}ling, Germany -- \\

(guneet.mutreja, ksenia.bittner)@dlr.de
}
% \address{**** (for review, affiliations must be rendered anonymous)}

% If the corresponding author is NOT the final author, always add a % space before the subsequent comma, i.e.
% first author name\textsuperscript{a,}\thanks{Corresponding author} , % second author name \textsuperscript{b}, etc.
% thanks to Niclas Borlin 05-05-2016
% information on the corresponding author should not be used any longer and has been commented out
% C. Heipke, Jan 03,2024

% the use of the information of commissions and working groups should not be used any longer and has been commented out
% C. Heipke, Sept. 20,2022
%\commission{XX, }{YY} %This field is optional. If filled, XX and YY should be replaced by adequate numbers. See https://www2.isprs.org/commissions/
%\workinggroup{XX/YY} %This field is optional.
%\icwg{}   %This field is optional.

% KAO: Use times symbol
\abstract{

Accurate classification of building roof types from aerial imagery is crucial for various remote sensing applications, including urban planning, disaster management, and infrastructure monitoring. However, this task is often hindered by the limited availability of labeled data for supervised learning approaches. To address this challenge, this paper investigates the effectiveness of self-supervised learning with EfficientNet architectures, known for their computational efficiency, for building roof type classification.
We propose a novel framework that incorporates a \gls{gl:CBAM} to enhance the feature extraction capabilities of EfficientNet. Furthermore, we explore the benefits of pretraining on a domain-specific dataset, the \gls{gl:AID}, compared to ImageNet pretraining. Our experimental results demonstrate the superiority of our approach. Employing \gls{gl:SimCLR} with EfficientNet-B3 and \gls{gl:CBAM} achieves a $95.5$\% accuracy on our validation set, matching the performance of state-of-the-art transformer-based models while utilizing significantly fewer parameters. We also provide a comprehensive evaluation on two challenging test sets, demonstrating the generalization capability of our method. Notably, our findings highlight the effectiveness of domain-specific pretraining, consistently leading to higher accuracy compared to models pretrained on the generic ImageNet dataset. Our work establishes EfficientNet-based self-supervised learning as a computationally efficient and highly effective approach for building roof type classification, particularly beneficial in scenarios with limited labeled data.
}


\maketitle

%\saythanks % added 28-02-2014 Markus Englich
\glsresetall
\section{Introduction}\label{INTRODUCTION}
 
% KAO: Sloppy spacing ensures non-overfull lines. Can be removed if this is not an issue.
\sloppy

\subsection{Background}\label{sec:Background}

The proliferation of remote sensing technologies has made aerial imagery readily available, opening up a wide range of applications for understanding and managing our built environment. Accurate and up-to-date information about buildings, including their roof structures, is crucial for urban planning, disaster response, infrastructure management, and the creation of detailed digital twins. For instance, precise 3D building models, which rely on accurate roof type classification, are essential for simulating urban environments, analyzing shadows, and generating realistic visualizations. Furthermore, knowing the distribution of different roof types is valuable for damage assessment after natural disasters and for energy modeling to optimize energy consumption and mitigate urban heat islands.

Despite its importance, obtaining sufficient labeled data for training robust roof type classification models remains a persistent challenge in remote sensing. Manually annotating aerial imagery to identify diverse roof structures is time-consuming, requires expert knowledge, and can be cost-prohibitive. This challenge is amplified by the global variability in roof architectures, ranging from simple flat roofs to complex combinations of gables, hips, and dormers. Additionally, dense urban environments often present difficulties in accurately delineating individual roofs due to closely spaced buildings and complex building shapes.

To address the limitations of relying on scarce labeled data, this paper explores the potential of self-supervised learning for building roof type classification. Self-supervised learning allows models to learn from readily available unlabeled data by leveraging inherent structures and patterns within the data itself to generate supervisory signals. This approach holds significant promise for remote sensing applications where labeled data is often limited. Specifically, we propose a novel framework that combines the computational efficiency of EfficientNet architectures, enhanced with a \gls{gl:CBAM}, and domain-specific pretraining on the \gls{gl:AID} to achieve accurate and efficient roof type classification, even with limited labeled examples.

\begin{figure}[t]
  \centering
  \setlength{\fboxsep}{0pt}
  \fbox{\includegraphics[width=0.95\linewidth]{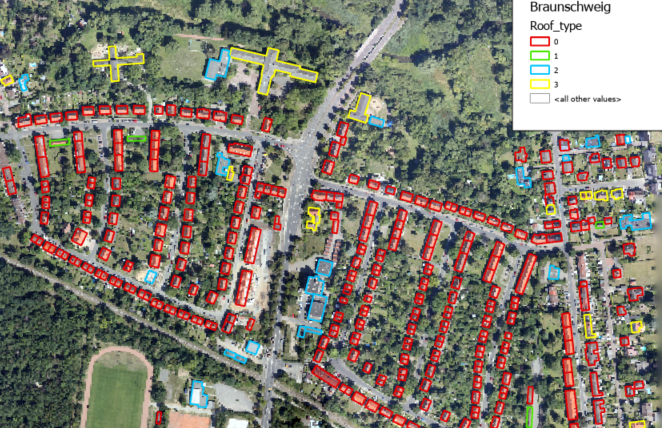}}

   \caption{Classification results of BEit model pretrained using Aerial Image Dataset on building roof types in the Braunschweig area.}
   \label{fig:braunschweig}
\end{figure}

% In Section~\ref{MANUSCRIPT} we present related work
%\newpage            
\subsection{Related work}\label{sec:Related works}

\textbf{Self-Supervised Learning}. The scarcity of labeled data in remote sensing has spurred the development of self-supervised learning methods, enabling models to learn representations from unlabeled data. These methods have gained significant traction due to their ability to leverage the abundance of unlabeled aerial imagery and reduce the reliance on costly manual annotations. A prominent category of self-supervised learning is contrastive learning, where models are trained to distinguish between augmented views of the same image (positive pairs) and different images (negative pairs). \cite{SimCLR}, \cite{MoCo}, and \cite{SWAV} are notable examples of contrastive learning methods that have shown success in learning useful representations for various downstream tasks, including image classification and object detection. These methods encourage models to learn invariant features that are robust to different viewpoints, lighting conditions, and other variations commonly encountered in remote sensing imagery. Other approaches leverage pretext tasks, where models are trained to solve a predefined task using the unlabeled data, as a form of self-supervision. Examples include \cite{Selfsupervised_one}, \cite{Selfsupervised_two}, or \cite{jigsaw}. While these pretext tasks might not be directly related to the final downstream task, they encourage models to learn meaningful representations that can be transferred and fine-tuned for specific applications.

More recently, transformer-based models have emerged as powerful tools for self-supervised learning in computer vision. \cite{mae}, inspired by the success of \cite{bert} in natural language processing, employ a masking strategy where portions of the input image are masked, and the model is trained to reconstruct the missing pixels. \cite{beit} and \cite{beitv2} leverage masked image modeling for pretraining, achieving state-of-the-art results on various image classification benchmarks. \cite{mocov3} combines contrastive and generative objectives by incorporating masked image modeling into the momentum contrastive (MoCo) framework. 

This paper investigates the effectiveness of several of these self-supervised learning approaches, including \gls{gl:SimCLR}, \gls{gl:SwAV}, and BEit, for the specific task of building roof type classification from aerial imagery.

\textbf{Building Roof Type Classification}. Classifying building roof types from aerial imagery is a crucial task in remote sensing, with applications in urban planning, 3D city modeling, disaster response, and energy modeling. However, developing accurate and robust methods for this task is challenging, primarily due to the difficulty of obtaining large, high-quality labeled datasets. Traditional approaches to roof type classification often relied on handcrafted features and rule-based systems \citep{rule}. However, these methods typically struggle to generalize to diverse roof structures and are sensitive to variations in image quality and scene complexity.

With the advent of deep learning, supervised learning approaches using convolutional neural networks (\gls{gl:CNN}) have become dominant in building roof type classification. Interestingly, \citet{roof} showed that even a shallow \gls{gl:CNN} architecture can achieve acceptable accuracy for roof type classification from very high-resolution orthophotos, particularly when dealing with sparse data. They compared their shallow \gls{gl:CNN} to fine-tuned versions of VGG-16 \citep{vgg}, EfficientNetB4 \citep{efficientnet}, and ResNet-50 \citep{Resnet50}, finding that while the shallow model had slightly lower overall accuracy, it still performed adequately for many applications. This suggests that computationally efficient \gls{gl:CNN} models can be effective for roof type classification. \citet{ubcv2} introduced the Urban Building Classification (UBC) V2-A Benchmark, a large-scale dataset for global building detection and fine-grained classification from satellite imagery, demonstrating the capabilities of \gls{gl:CNN}s for this task. While promising, these supervised methods require substantial amounts of labeled data for training. 

Despite the advancements made with deep learning, the reliance on large labeled datasets remains a significant bottleneck for supervised approaches. Creating such datasets is labor-intensive, time-consuming, and expensive, especially when considering the global diversity of roof structures and the challenges posed by dense urban environments. This limitation motivates the exploration of self-supervised learning methods, which can learn effective representations from unlabeled data, as investigated in this paper.

\textbf{EfficientNets for Object Recognition}. EfficientNets have evolved as a compelling alternative to more resource-intensive architectures like ResNets \citep{Resnet50}. Introduced by \citet{efficientnet}, EfficientNets offer a balanced trade-off between computational complexity and performance. Their unique scaling method optimizes depth, width, and resolution, thereby enabling performance gains with fewer parameters and less computational cost. Given the specific challenges posed by aerial object recognition, wherein the availability of computational resources can be a limiting factor, EfficientNets stand as a viable alternative to traditional architectures. Recent works have started to examine the integration of EfficientNets within self-supervised frameworks \citep{self_efficientnets}, which has influenced our choice for this study.

\textbf{Domain-Specific Pretraining}. While pretraining models on large, diverse datasets like ImageNet \citep{imagenet} has been a common practice, there is a growing realization that domain-specific pretraining can offer distinct advantages \citep{domain_pretraining}. This is particularly true in specialized fields such as aerial object recognition, where the statistical properties of images can differ considerably from generic datasets. By utilizing a domain-specific dataset for pretraining - like the \gls{gl:AID} \citep{AID} in our case - the model is better aligned with the characteristics inherent to aerial imagery, thereby potentially improving performance and generalizability.

While previous research has explored self-supervised learning, EfficientNets, and domain-specific pretraining separately, this paper presents a novel framework that integrates these advancements to address the challenges of accurate and efficient building roof type classification from aerial imagery, especially in scenarios with limited labeled data.

\section{Methodology and Experiments}\label{sec:METHOD}

\begin{figure*}[t] % 't' ensures the image stays at the top of the page
  \centering
  \includegraphics[width=\linewidth]{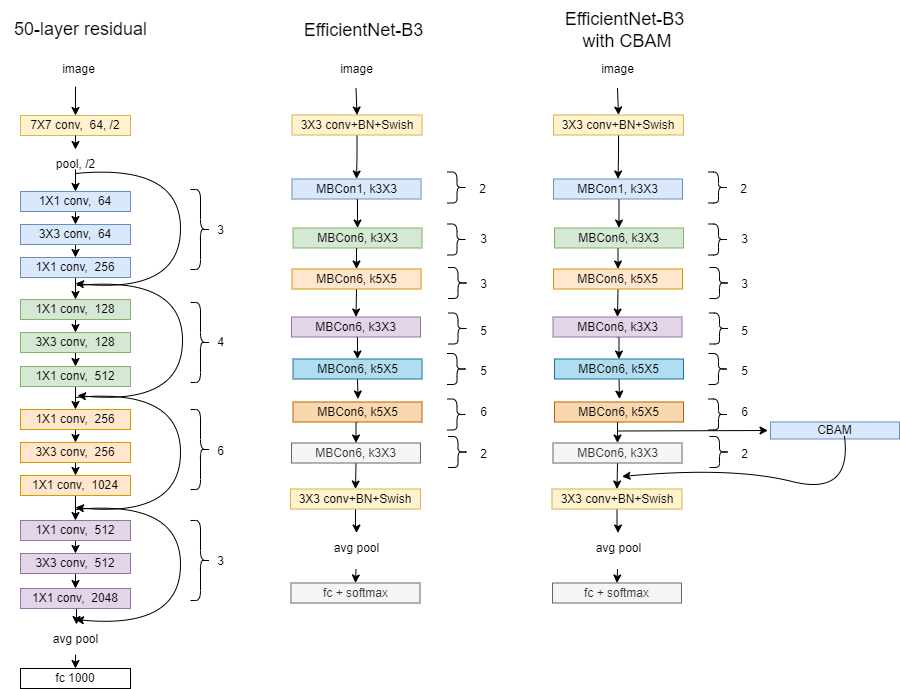}
  \caption{Visual representation of representative architecture configurations employed in the study: ResNet50, EfficientNetB3, and EfficientNetB3 enhanced with \gls{gl:CBAM}. These architectures serve as examples of the diverse backbones explored, ranging from ResNet34 to ResNet50 and EfficientNetB0 to B3 with and without \gls{gl:CBAM}.}
  \label{fig:backbones}
\end{figure*}

This section presents the methodology employed for classifying building roof types from aerial imagery using self-supervised learning. We begin with an overview of our two-phase approach: pretraining and linear evaluation. We then provide a detailed description of the datasets used for both phases, followed by explanations of the self-supervised learning methods and the architecture choices employed for pretraining. Finally, we outline the linear evaluation protocol used to assess the performance of our approach on the roof type classification task.

\subsection{Pretraining with Self-Supervised Learning}\label{sec:Pretraining with Self-Supervised Learning}

\textbf{\gls{gl:AID} for Pretraining}. The \gls{gl:AID} \citep{AID} is a large-scale dataset explicitly designed for aerial scene classification, making it suitable for pretraining models on a diverse range of aerial imagery. It consists of 10,000 images categorized into 30 different scene classes, including airports, beaches, bridges, and various urban and rural landscapes. The images in \gls{gl:AID} have a spatial resolution of $600\times600$ pixels. No additional preprocessing was applied to the images beyond the standard normalization procedures provided by the \emph{PyTorch} library.
We selected \gls{gl:AID} for pretraining because it contains a wide variety of aerial scenes, enabling our models to learn generic features and representations relevant to the domain of aerial imagery. We hypothesize that pretraining on \gls{gl:AID}, a domain-specific dataset, will be more beneficial for the downstream task of roof type classification compared to pretraining on a more general-purpose dataset like ImageNet.

\textbf{Backbone Architectures.} For this study, we chose EfficientNets \citep{efficientnet} as our backbone architectures for their computational efficiency and high performance in image recognition tasks. EfficientNets are a family of convolutional neural networks that achieve state-of-the-art accuracy with significantly fewer parameters and computational resources compared to traditional \gls{gl:CNN}s like ResNets \citep{Resnet50}. Their efficiency is particularly important for processing large-scale aerial imagery datasets, which are often computationally demanding. 
We evaluated EfficientNet models ranging from B0 to B3, offering a trade-off between model complexity and accuracy. For comparison, we also included ResNet architectures (ResNet34 and ResNet50) in our experiments. \Cref{fig:backbones} provides a visual illustration of the ResNet-50 architecture, EfficientNet-B3 architecture and its variant enhanced with the \gls{gl:CBAM}, which will be described in the next section.
\begin{table*}[t]
  \centering
  \small
  \setlength{\tabcolsep}{1.8pt}
  \resizebox{\textwidth}{!}{ % Resize the table to fit within the text width
    \begin{tabular}{@{}lcccccccccc@{}}
      \toprule
      \multirow{2}{*}{Method} & \multicolumn{2}{c}{ResNet} & \multicolumn{4}{c}{EfficientNet} & \multicolumn{4}{c}{EfficientNet+CBAM} \\
      \cmidrule(lr){2-3} \cmidrule(lr){4-7} \cmidrule(lr){8-11}
      & 34 & 50 & B0 & B1 & B2 & B3 & B0 & B1 & B2 & B3 \\
      \midrule
      DenseCL & \( \textbf{81.8} \pm 0.6 \) & \( 72.1 \pm 0.4 \) & \( 74.9 \pm 1.9 \) & \( 75.4 \pm 0.2 \) & \( 72.2 \pm 0.7 \) & \( 66.1 \pm 0.8 \) & \( 76.9 \pm 1.01 \) & \( 71.5 \pm 0.3 \) & \( 70.8 \pm 3.01 \) & \( 66.79 \pm 2.6 \) \\
      MoCoV3 & \( 88.8 \pm 0.39 \) & \( \textbf{93.7} \pm 0.2 \) & \( 83.3 \pm 3.1 \) & \( 72.6 \pm 3.2 \) & \( 77.9 \pm 4.8 \) & \( 76.6 \pm 0.85 \) & \( 79.9 \pm 0.6 \) & \( 77.7 \pm 4.6 \) & \( 75.4 \pm 3.3 \) & \( 76.7 \pm 3.6 \) \\
      \gls{gl:SimCLR} & \( 83.6 \pm 0.2 \) & \( 86.8 \pm 0.1 \) & \( 94.9 \pm 0.1 \) & \( 94.6 \pm 0.2 \) & \( 94.8 \pm 0.1 \) & \( 94.7 \pm 0.5 \) & \( 95.1 \pm 0.26 \) & \( \textbf{95.5} \pm 0.07 \) & \( 94.9 \pm 0.1 \) & \( 94.7 \pm 0.12 \) \\
      \gls{gl:SwAV} & \( 72.5 \pm 3.8 \) & \( 72.1 \pm 4.2 \) & \( 50.4 \pm 3.94 \) & \( 73.4 \pm 0.58 \) & \( 72.2 \pm 0.43 \) & \( 72.0 \pm 0.89 \) & \( 76.8 \pm 0.5 \) & \( \textbf{78.6} \pm 0.2 \) & \( 75.7 \pm 0.2 \) & \( 75.8 \pm 0.3 \) \\
      \bottomrule
    \end{tabular}
  }
  \caption{Classification Accuracies for Building Roof Types Using \gls{gl:CNN}-Based Self-Supervised Models pretrained on AID: A Comparison Across ResNet, EfficientNet, and \gls{gl:CBAM}-Enhanced EfficientNet Encoders. The best performing models from each of the methods are highlighted.}
  \label{tab:results_conv}
\end{table*}

\textbf{Attention Mechanism.} To further enhance the feature extraction capabilities of our EfficientNet models, we incorporate the \gls{gl:CBAM} \citep{CBAM}. \gls{gl:CBAM} is a lightweight attention mechanism that can be seamlessly integrated into \gls{gl:CNN} architectures. It refines the feature maps by selectively attending to important spatial and channel-wise information.
In our approach, we append \gls{gl:CBAM} to the end of the EfficientNet backbone before the classification head. By integrating \gls{gl:CBAM}, we aim to improve the ability of the model to focus on the most relevant features for roof type classification, potentially leading to higher accuracy.

\textbf{Self-Supervised Learning Methods.} We explore a variety of self-supervised learning methods to pretrain our EfficientNet models. These methods can be broadly categorized into two groups: contrastive learning methods and transformer-based methods.

\begin{figure}[!t]
  \centering
  \captionsetup[subfigure]{font=LARGE}
  \resizebox{\linewidth}{!}{
    \begin{tabular}{cc}  % 2 columns, 2 rows
      % First row, first column
      \begin{subfigure}[t]{0.45\textwidth}
        \centering
        \includegraphics[height=6cm, width=\linewidth]{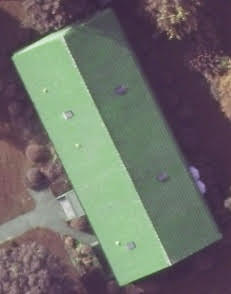}
        \caption{Gable roof}
      \end{subfigure} &
      % First row, second column
      \begin{subfigure}[t]{0.45\textwidth}
        \centering
        \includegraphics[height=6cm, width=\linewidth]{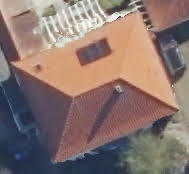}
        \caption{Hip roof}
      \end{subfigure} \\
      % Second row, first column
      \begin{subfigure}[t]{0.45\textwidth}
        \centering
        \includegraphics[height=6cm, width=\linewidth]{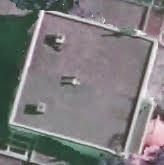}
        \caption{Flat roof}
      \end{subfigure} &
      % Second row, second column
      \begin{subfigure}[t]{0.45\textwidth}
        \centering
        \includegraphics[height=6cm, width=\linewidth]{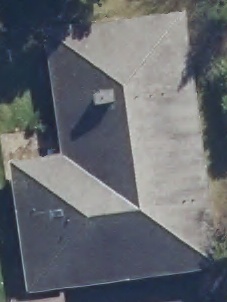}
        \caption{Complex roof}
      \end{subfigure}
    \end{tabular}
  }
  \caption{Sample images from the dataset illustrating the four roof types: Gable, Hip, Flat, and Complex.}
  \label{fig:roofs}
\end{figure}

\begin{itemize}
    \item Contrastive methods: \gls{gl:SimCLR}, MoCoV3, \gls{gl:SwAV} employ contrastive learning, where models learn to distinguish between augmented views of the same image (positive pairs) and different images (negative pairs). For these methods, we employed the LARS optimizer \citep{LARS} with a learning rate initially determined using a linear scaling rule. The learning rate increased during the first 10 epochs and decayed following a cosine schedule. A weight decay of $1e-6$ was applied. For DenseCL, we used Stochastic Gradient Descent (SGD) with a learning rate of $0.03$, a momentum of $0.9$, and a weight decay of $1e-4$.
    \item Transformer-based methods: For BEit, BEitV2, \gls{gl:MAE}, we used the AdamW optimizer \citep{adamw} with a weight decay of $0.1$ and an initial learning rate of $1e-4$ with a warmup period of $40$ epochs.
\end{itemize}

We pretrain our models using each of these methods for $300$ epochs on the \gls{gl:AID} dataset. The specific hyperparameters used for each method, including optimizer, learning rate, batch size, and data augmentation techniques, will be provided in the following section on Linear Evaluation Protocol. All experiments were conducted using the MMPreTrain library \citep{2023mmpretrain} on an NVIDIA Titan RTX GPU with $24$GB of memory.

\subsection{Linear Evaluation}\label{sec:Linear Evaluation Protocolg}

\textbf{Dataset and labeling.} To evaluate the effectiveness of the pretrained EfficientNet models for building roof type classification, we perform a linear evaluation. This involves fine-tuning the pretrained models on a labeled dataset of aerial images and assessing their performance on the roof type classification task. We utilized building roof vectorization dataset \citep{roofdataset}, originally lacking roof type information. We manually annotated the images to classify them into four roof types: Gable, Hip, Flat, and Complex, as illustrated in \Cref{fig:roofs}. The images are RGB orthophotos with a spatial resolution of $0.1$ meters per pixel. The labeled dataset is divided into a training set of $7500$ images and a validation set of $765$ images.

\begin{table}[t]
  \centering
  \begin{tabular}{@{}lc@{}}
    \toprule
    Method & ViTBase \\
    \midrule
    BEit & \( \textbf{95.1} \pm 0.2 \) \\
    BEitV2 & \( 93.7 \pm 0.1 \) \\
    \gls{gl:MAE} & \( 77.1 \pm 0.3 \) \\
    MoCoV3 & \( 73.3 \pm 4.9 \)\\
    \bottomrule
  \end{tabular}
  \caption{Classification Accuracies for Building Roof Types Using Transformer-Based Self-Supervised Models pretrained on AID: A Comparison on ViTBase Encoder.}
  \label{tab:results_vit}
\end{table}

\textbf{Data Augmentation and Training Details.} For data augmentation, random resize crops and horizontal flips were applied. The models were then linearly evaluated for $100$ epochs on this dataset using the following optimization strategies:
\begin{itemize}
\item For \gls{gl:SimCLR}, \gls{gl:SwAV}, and MoCoV3, the LARS optimizer was used with a learning rate of $0.6$ and a momentum of $0.9$.
\item For DenseCL, we used SGD with a learning rate of $0.1$, a momentum of $0.9$, and a weight decay of $1e-4$.
\item For BEit and \gls{gl:MAE}, we used SGD with a learning rate of $0.1$, a momentum of $0.9$, and a weight decay of $1e-3$.
\end{itemize}

The learning rates for all models were decayed according to cosine schedule.

\begin{figure*}[ht!]
    \centering
    \tabcolsep=0.05cm
    \begin{tabular}{cccc}
        \begin{subfigure}[t]{0.24\linewidth}
            \centering
            \adjustbox{valign=t}{\includegraphics[height=4cm,width=\linewidth]{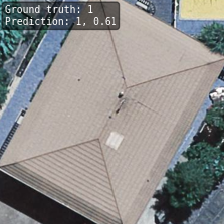}}
            \caption{Hip roof}\label{subfig:ab}
        \end{subfigure}%
        \hspace{0.01\textwidth}
        \begin{subfigure}[t]{0.24\linewidth}
            \centering
            \adjustbox{valign=t}{\includegraphics[height=4cm,width=\linewidth]{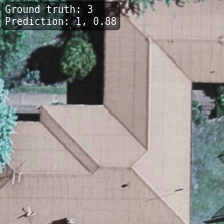}}
            \caption{Complex roof}
        \end{subfigure}%
        \hspace{0.01\textwidth}
        \begin{subfigure}[t]{0.24\linewidth}
            \centering
            \adjustbox{valign=t}{\includegraphics[height=4cm,width=\linewidth]{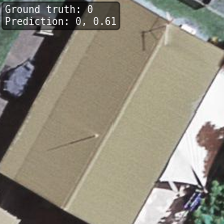}}
            \caption{Gable roof}
        \end{subfigure}%
        \hspace{0.01\textwidth}
        \begin{subfigure}[t]{0.24\linewidth}
            \centering
            \adjustbox{valign=t}{\includegraphics[height=4cm,width=\linewidth]{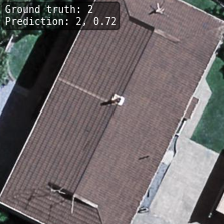}}
            \caption{Flat roof}
        \end{subfigure}\\[1ex]
        \begin{subfigure}[t]{0.24\linewidth}
            \centering
            \adjustbox{valign=t}{\includegraphics[height=4cm,width=\linewidth]{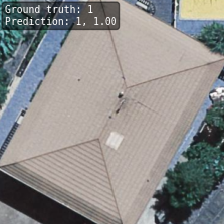}}
            \caption{Hip roof}
        \end{subfigure}%
        \hspace{0.01\textwidth}
        \begin{subfigure}[t]{0.24\linewidth}
            \centering
            \adjustbox{valign=t}{\includegraphics[height=4cm,width=\linewidth]{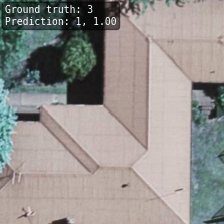}}
            \caption{Complex roof}
        \end{subfigure}%
        \hspace{0.01\textwidth}
        \begin{subfigure}[t]{0.24\linewidth}
            \centering
            \adjustbox{valign=t}{\includegraphics[height=4cm,width=\linewidth]{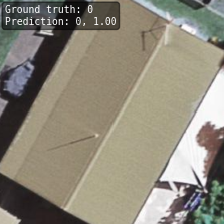}}
            \caption{Gable roof}
        \end{subfigure}%
        \hspace{0.01\textwidth}
        \begin{subfigure}[t]{0.24\linewidth}
            \centering
            \adjustbox{valign=t}{\includegraphics[height=4cm,width=\linewidth]{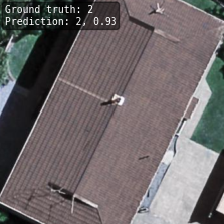}}
            \caption{Flat roof}
        \end{subfigure}\\[1ex]
    \end{tabular}
    \caption{Performance visualization: BEit (Row 1) and \gls{gl:SimCLR} (Row 2) on a manually labeled Roof Graph subsets, pretrained on AID.} 
    \label{fig:results}
\end{figure*}

\begin{table*}[ht!]
  \centering
  \small
  \setlength{\tabcolsep}{1.5pt}
  \resizebox{\textwidth}{!}{ % Resize the table to fit within the text width
      \begin{tabular}{@{}lcccccccccc@{}}
        \toprule
        \multirow{2}{*}{Input} & \multicolumn{2}{c}{ResNet} & \multicolumn{4}{c}{EfficientNet} & \multicolumn{4}{c}{EfficientNet+CBAM} \\
        \cmidrule(lr){2-3} \cmidrule(lr){4-7} \cmidrule(lr){8-11}
        image & 34 & 50 & B0 & B1 & B2 & B3 & B0 & B1 & B2 & B3 \\
        \midrule
         \includegraphics[width=0.08\textwidth, height=0.05\textheight]{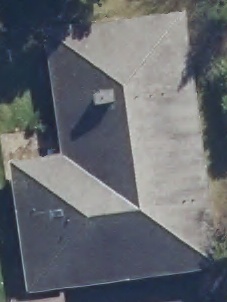} & \includegraphics[width=0.08\textwidth, height=0.05\textheight]{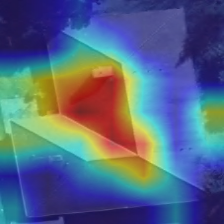} & \includegraphics[width=0.08\textwidth, height=0.05\textheight]{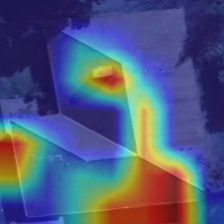} & \includegraphics[width=0.08\textwidth, height=0.05\textheight]{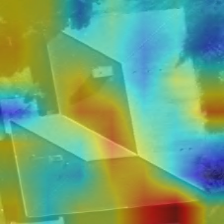} & \includegraphics[width=0.08\textwidth, height=0.05\textheight]{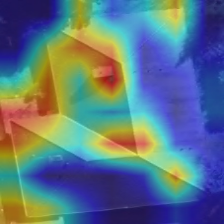} & \includegraphics[width=0.08\textwidth, height=0.05\textheight]{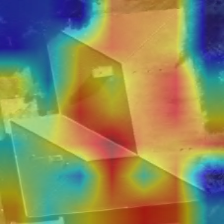} & \includegraphics[width=0.08\textwidth, height=0.05\textheight]{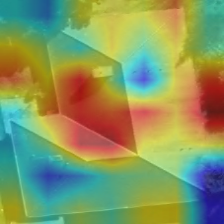} & \includegraphics[width=0.08\textwidth, height=0.05\textheight]{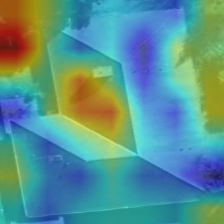} & \includegraphics[width=0.08\textwidth, height=0.05\textheight]{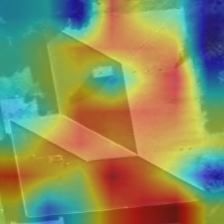} & \includegraphics[width=0.08\textwidth, height=0.05\textheight]{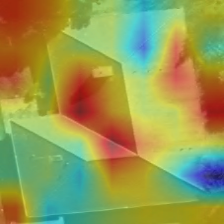} &
        \includegraphics[width=0.08\textwidth, height=0.05\textheight]{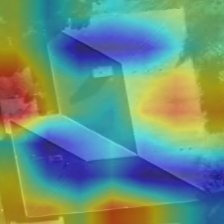} \\
        \includegraphics[width=0.08\textwidth, height=0.05\textheight]{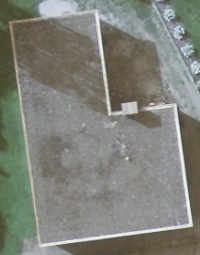} & \includegraphics[width=0.08\textwidth, height=0.05\textheight]{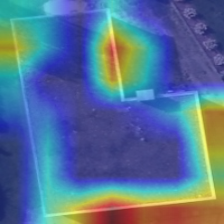} & \includegraphics[width=0.08\textwidth, height=0.05\textheight]{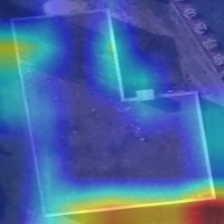} & \includegraphics[width=0.08\textwidth, height=0.05\textheight]{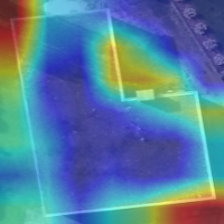} & \includegraphics[width=0.08\textwidth, height=0.05\textheight]{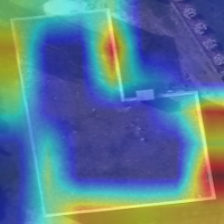} & \includegraphics[width=0.08\textwidth, height=0.05\textheight]{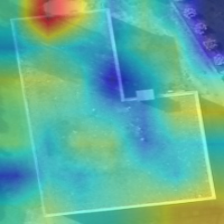} & \includegraphics[width=0.08\textwidth, height=0.05\textheight]{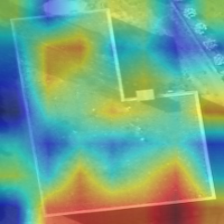} & \includegraphics[width=0.08\textwidth, height=0.05\textheight]{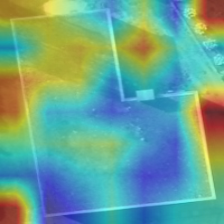} & \includegraphics[width=0.08\textwidth, height=0.05\textheight]{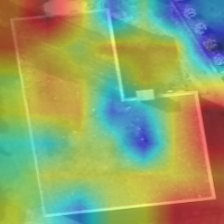} & \includegraphics[width=0.08\textwidth, height=0.05\textheight]{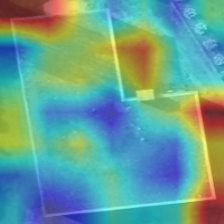} &
        \includegraphics[width=0.08\textwidth, height=0.05\textheight]{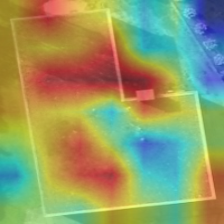} \\
        \includegraphics[width=0.08\textwidth, height=0.05\textheight]{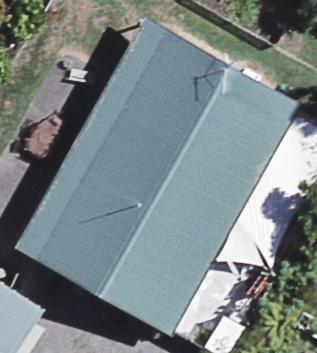} & \includegraphics[width=0.08\textwidth, height=0.05\textheight]{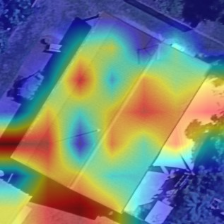} & \includegraphics[width=0.08\textwidth, height=0.05\textheight]{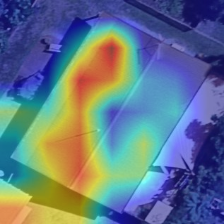} & \includegraphics[width=0.08\textwidth, height=0.05\textheight]{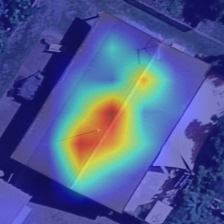} & \includegraphics[width=0.08\textwidth, height=0.05\textheight]{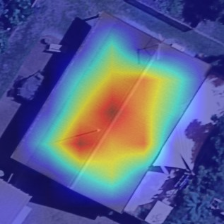} & \includegraphics[width=0.08\textwidth, height=0.05\textheight]{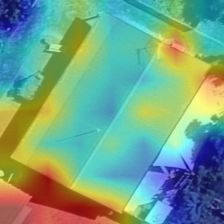} & \includegraphics[width=0.08\textwidth, height=0.05\textheight]{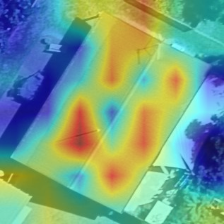} & \includegraphics[width=0.08\textwidth, height=0.05\textheight]{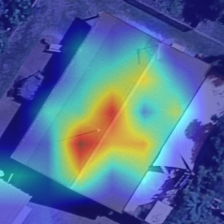} & \includegraphics[width=0.08\textwidth, height=0.05\textheight]{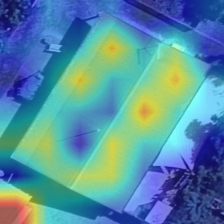} & \includegraphics[width=0.08\textwidth, height=0.05\textheight]{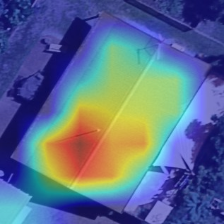} &
        \includegraphics[width=0.08\textwidth, height=0.05\textheight]{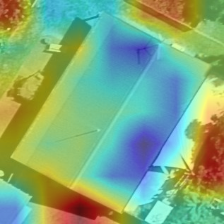} \\
        \includegraphics[width=0.08\textwidth, height=0.05\textheight]{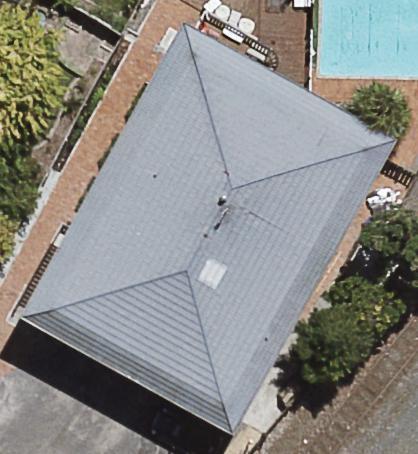} & \includegraphics[width=0.08\textwidth, height=0.05\textheight]{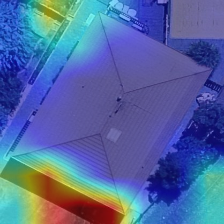} & \includegraphics[width=0.08\textwidth, height=0.05\textheight]{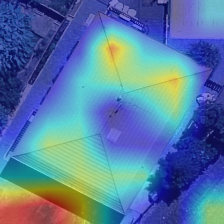} & \includegraphics[width=0.08\textwidth, height=0.05\textheight]{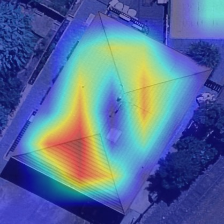} & \includegraphics[width=0.08\textwidth, height=0.05\textheight]{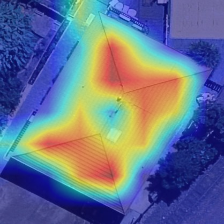} & \includegraphics[width=0.08\textwidth, height=0.05\textheight]{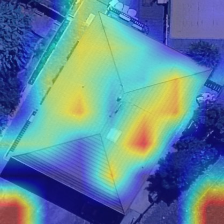} & \includegraphics[width=0.08\textwidth, height=0.05\textheight]{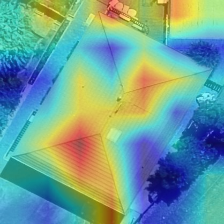} & \includegraphics[width=0.08\textwidth, height=0.05\textheight]{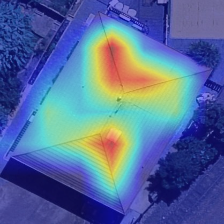} & \includegraphics[width=0.08\textwidth, height=0.05\textheight]{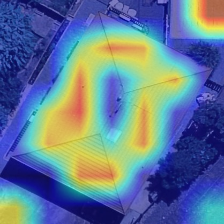} & \includegraphics[width=0.08\textwidth, height=0.05\textheight]{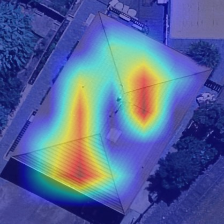} &
        \includegraphics[width=0.08\textwidth, height=0.05\textheight]{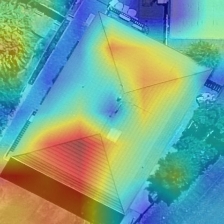} \\
        \bottomrule
      \end{tabular}%
  }
  \caption{Class Activation Maps generated by the \gls{gl:SimCLR} model pretrained on the Aerial Image Dataset. Each row represents a different roof category: the first row contains Complex, the second Flat, the third Gable, and the fourth Hip.}
  \label{tab:simclr}
\end{table*}

\section{Results and Discussion}
\label{sec:resulta}

\subsection{Performance of Convolutional Models}

We evaluated various convolutional models, including ResNet and EfficientNet variants, with and without the \gls{gl:CBAM}. As shown in Table \ref{tab:results_conv}, incorporating \gls{gl:CBAM} consistently improved the performance of the EfficientNet models. For instance, DenseCL with EfficientNet-B0 and CBAM achieved an accuracy of $76.9$\%, compared to $74.9$\% for the same model without \gls{gl:CBAM}. This trend suggests that \gls{gl:CBAM}'s selective attention mechanism effectively enhances feature extraction for building roof type classification.

\subsection{Comparative Performance: \gls{gl:SimCLR} and Transformer-Based Models}
Notably, the \gls{gl:SimCLR} method combined with EfficientNet and \gls{gl:CBAM} consistently outperformed other architectures, even exceeding the accuracy of transformer-based models like BEit and BEitV2 in certain cases, as shown in \Cref{tab:results_conv} and \Cref{tab:results_vit}. For example, \gls{gl:SimCLR} with EfficientNet-B1 and \gls{gl:CBAM} achieved a remarkable accuracy of $95.5$\%, representing a substantial $12$\% improvement over its ResNet-pretrained counterpart. This result underscores the effectiveness and efficiency of combining EfficientNet backbones with \gls{gl:CBAM} for self-supervised learning tasks in building roof type classification.

The Class Activation Maps generated by \gls{gl:SimCLR} dramatically validate our observation about its exceptional performance when combined with EfficientNet and \gls{gl:CBAM} as shown in \Cref{tab:simclr}. ResNet 34 and 50 exhibit lower activations compared to all variants of EfficientNet. Most notably, the \gls{gl:CBAM}-enhanced EfficientNets show the highest activations, supporting our claim that \gls{gl:SimCLR} optimized with EfficientNet and \gls{gl:CBAM} exceeds its ResNet versions in performance.

\subsection{Impact of Pretraining Datasets: ImageNet-1k vs. Domain-Specific}
While models pretrained on the ImageNet-1k dataset generally demonstrated good performance, as shown in Table \ref{tab:results_IN1K}, certain methods exhibited significant improvements when pretrained on the domain-specific \gls{gl:AID}. \gls{gl:SimCLR} and \gls{gl:MAE}, in particular, showed substantial gains with domain-specific pretraining. For instance, \gls{gl:SimCLR}'s accuracy increased from $80.78$\% with ImageNet-1k pretraining to $83.6$\% with \gls{gl:AID} pretraining for ResNet50 backbone. This finding underscores the importance of leveraging domain-specific data for pretraining, especially in specialized applications like building roof type classification.

\subsection{Generalization to Test Datasets}

To assess the generalizability of our approach, we evaluated the two best-performing models, \gls{gl:SimCLR} (with EfficientNet-B3 + CBAM) and BEit, both pretrained on the AID dataset, on two independent test sets. 
The first test set, the Braunschweig dataset, consists of approximately $350$ building polygons extracted from the Braunschweig city dataset (acquired from CrowdAI) and manually labeled for roof type information. The images are 3-band RGB orthophotos with a spatial resolution of 0.3 meters from our flight campaign. The second test set, the Roof Graph subset, comprises $300$ manually labeled images randomly selected from the Roof Graph dataset \citep{roofdataset}, which covers the city of Detmold, Germany. The images are 3-band RGB orthophotos with a resolution of 0.1 meters per pixel.

\begin{table}
  \centering
  \begin{tabular}{@{}lc@{}}
    \toprule
    Method & ViTBase/ResNet50  \\
    \midrule
    BEit & \( 96.4 \pm 0.7 \) \\
    BEitV2 & \( 96.8 \pm 0.3 \) \\
    DenseCL & \( 94.1 \pm 0.4 \) \\
    \gls{gl:MAE} & \( 77.1 \pm 2.0 \) \\
    MoCoV3 ResNet & \( 95.8 \pm 0.08 \) \\
    MoCoV3 ViT & \( 92.9 \pm 0.2 \) \\
    \gls{gl:SimCLR} & \( 80.78 \pm 0.2 \) \\
    \gls{gl:SwAV} & \( 87.9 \pm 0.3 \) \\
    \bottomrule
  \end{tabular}
  \caption{Classification Accuracies for Building Roof Types Using Transformer-Based and \gls{gl:CNN}-Based Self-Supervised Models pretrained on ImageNet-1K dataset: A Comparison on ViTBase and ResNet50 Encoder.}
  \label{tab:results_IN1K}
\end{table}

\begin{table}
  \centering
  \vspace{0.2cm}
  \begin{tabular}{ccc}
    \toprule
    Method & \makecell{Braunschweig \\ dataset} & \makecell{Roof graph \\\ Subset} \\
    \midrule
    BEit & $90.7$ & $81$\\
    \gls{gl:SimCLR} & $87.2$ & $75.2$\\
    \bottomrule
  \end{tabular}
  \caption{Classification Accuracies of \gls{gl:SimCLR} (EfficientNet-B1 + \gls{gl:CBAM}) and BEit model pretrained on AID dataset on test datasets.}
  \label{tab:results_test}
\end{table}

As shown in Table \ref{tab:results_test}, BEit achieved higher accuracies ($90.7$\% and $81$\%) on both the Braunschweig and Roof Graph subsets, respectively, compared to \gls{gl:SimCLR} ($87.2$\% and $75.2$\%). These results demonstrate the robust performance of both methods on unseen data, supporting the generalization capabilities of our self-supervised pretraining approach. Figures \ref{fig:results} and \ref{fig:braunschweig} provide visual examples of the classification results on these test sets.

\subsection{Computational Efficiency}
\label{sec:computation}
In practical remote sensing applications, computational efficiency is crucial, especially when dealing with large-scale aerial imagery datasets. Table \ref{tab:Computations} compares the computational requirements of ResNet50, EfficientNet-B3, and EfficientNet-B3 with \gls{gl:CBAM}. EfficientNet-B3 demonstrates significant advantages, requiring substantially fewer parameters ($13.255$M vs. $27.969$M) and FLOPS ($0.992$G vs. $4.109$G) compared to ResNet50 while achieving comparable or even better accuracy. The addition of \gls{gl:CBAM} only marginally increases the computational cost of EfficientNet-B3. These results highlight EfficientNet's efficiency, making it highly suitable for large-scale aerial imagery processing tasks where resource constraints and processing time are critical considerations.

\begin{table}[htbp]
  \centering
  \begin{tabularx}{\columnwidth}{|l|X|X|X|}
    \hline
    Arch. & Params
    (M) & FLOPS
    (G) & Avg. iter. time
    (s/iter) \\
    \hline\hline
    ResNet50 & $27.969$ & $4.109$ & $0.967$ \\
    EfficientNetB3 & $13.255$ & $0.992$ & $0.3588$ \\
    EfficientNetB3 + \gls{gl:CBAM} & $13.274$ & $0.992$ & $0.3597$ \\
    \hline
  \end{tabularx}
  \caption{\textbf{Compute requirements.} We present the total number of parameters, FLOPS (floating point operations per second), and the average iteration time required for the three architectures.}
  \label{tab:Computations}
\end{table}

\glsresetall
\section{Conclusion}

Our results demonstrate the significant potential of self-supervised learning for building roof type classification from aerial imagery. The combination of EfficientNet architectures and the \gls{gl:CBAM} proved particularly effective. Incorporating \gls{gl:CBAM} consistently enhanced the performance of EfficientNet models, highlighting the benefits of selective attention for this task. Moreover, EfficientNet consistently outperformed ResNet, achieving higher accuracy while requiring substantially fewer parameters and computational resources as demonstrated in \ref{sec:computation}.

Among the self-supervised methods evaluated, \gls{gl:SimCLR}, when combined with EfficientNet-B3 and \gls{gl:CBAM}, achieved the highest classification accuracy. This method not only outperformed other \gls{gl:CNN}-based approaches but also demonstrated comparable or superior performance to transformer-based models in certain cases.

Our findings strongly emphasize the advantage of domain-specific pretraining. Models pretrained on the \gls{gl:AID} consistently achieved higher accuracy than those pretrained on the general-purpose ImageNet-1k dataset. This highlights the importance of leveraging datasets that reflect the unique characteristics of aerial imagery for specialized tasks like building roof type classification.

The promising results obtained in this study lay the foundation for further exploration of self-supervised learning in remote sensing. Future work could investigate the application of these methods to more complex roof type categories or explore the integration of other advanced architectures and attention mechanisms. Additionally, expanding the domain-specific pretraining dataset with more diverse aerial imagery could further improve performance and generalization capabilities.

\bibliography{ISPRSguidelines_authors}

\end{document}